\documentclass[11pt,a4paper]{article}
\usepackage[hyperref]{acl2017}
\usepackage{times}
\usepackage{latexsym}
\usepackage{enumitem}

\PassOptionsToPackage{hyphens}{url}\usepackage{hyperref}
\usepackage{booktabs}
\usepackage{multirow}

\usepackage{listings}
\usepackage{amsmath}
\usepackage{tabulary}
\usepackage{tabularx}
\usepackage{graphicx}
\usepackage{color}
\graphicspath{ {Images/} }

\aclfinalcopy 

\title{A Supervised Approach to Extractive Summarisation of Scientific Papers}

\author{Ed Collins \and Isabelle Augenstein \and Sebastian Riedel \\
  Department of Computer Science,\\ University College London (UCL), UK \\
  {\tt \{edward.collins.13|i.augenstein|s.riedel\}@ucl.ac.uk}}

\date{}

\begin{document}
\maketitle
\begin{abstract}
Automatic summarisation is a popular approach to reduce a document to its main arguments. Recent research in the area has focused on neural approaches to summarisation, which can be very data-hungry. However, few large datasets exist and none for the traditionally popular domain of scientific publications, which opens up challenging research avenues centered on encoding large, complex documents. In this paper, we introduce a new dataset for summarisation of computer science publications by exploiting a large resource of author provided summaries and show straightforward ways of extending it further. We develop models on the dataset making use of both neural sentence encoding and traditionally used summarisation features and show that models which encode sentences as well as their local and global context perform best, significantly outperforming well-established baseline methods. 
\end{abstract}

\section{Introduction}

Automatic summarisation is the task of reducing a document to its main points. There are two streams of summarisation approaches: \textit{extractive summarisation}, which copies parts of a document (often whole sentences) to form a summary, and \textit{abstractive summarisation}, which reads a document and then generates a summary from it, which can contain phrases not appearing in the document.
Abstractive summarisation is the more difficult task, but useful for domains where sentences taken out of context are not a good basis for forming a grammatical and coherent summary, like novels.

Here, we are concerned with summarising scientific publications. Since scientific publications are a technical domain with fairly regular and explicit language, we opt for the task of \textit{extractive summarisation}. Although there has been work on summarisation of scientific publications before, existing datasets are very small, consisting of tens of documents~\cite{kupiec1995trainable,Visser2009}. 
Such small datasets are not sufficient to learn supervised summarisation models relying on neural methods for sentence and document encoding, usually trained on many thousands of documents~\cite{rush-chopra-weston:2015:EMNLP,cheng-lapata:2016:P16-1,chopra-auli-rush:2016:N16-1,see2017get}.

In this paper, we introduce a dataset for automatic summarisation of computer science publications which can be used for both abstractive and extractive summarisation. It consists of more than 10k documents and can easily be extended automatically to an additional 26 domains. The dataset is created by exploiting an existing resource, ScienceDirect,\footnote{\url{http://www.sciencedirect.com/}} where many journals require authors to submit highlight statements along with their manuscripts. Using such highlight statements as gold statements has been proven a good gold standard for news documents~\cite{nallapati2016summarunner}.
This new dataset offers many exciting research challenges, such how best to encode very large technical documents, which are largely ignored by current research.

\setlength{\tabcolsep}{0.3em}
\begin{table}[t]
\fontsize{8}{8}\selectfont
\begin{center}
\begin{tabularx}{\linewidth}{X}
\toprule
\textbf{Paper Title} Statistical estimation of the names of HTTPS servers with domain name graphs \\ \midrule
\textbf{Highlights} we present the domain name graph (DNG), which is a formal expression that can keep track of cname chains and characterize the dynamic and diverse nature of DNS mechanisms and deployments. We develop a framework called service-flow map (sfmap) that works on top of the DNG.sfmap estimates the hostname of an HTTPS server when given a pair of client and server IP addresses. It can statistically estimate the hostname even when associating DNS queries are unobserved due to caching mechanisms, etc through extensive analysis using real packet traces, we demonstrate that the sfmap framework establishes good estimation accuracies and can outperform the state-of-the art technique called dn-hunter. We also identify the optimized setting of the sfmap framework. The experiment results suggest that the success of the sfmap lies in the fact that it can complement incomplete DNS information by leveraging the graph structure. To cope with large-scale measurement data, we introduce techniques to make the sfmap framework scalable. We validate the effectiveness of the approach using large-scale traffic data collected at a gateway point of internet access links .\\ \midrule
\textbf{Summary Statements Highlighted in Context from Section of Main Text}
Contributions: in this work, we present a novel methodology that aims to infer the hostnames of HTTPS flows, given the three research challenges shown above. The key contributions of this work are summarized as follows. {\color{red} We present the domain name graph (DNG), which is a formal expression that can keep track of cname chains (challenge 1) and characterize the dynamic and diverse nature of DNS mechanisms and deployments (challenge 3). We develop a framework called service-flow map (sfmap) that works on top of the DNG. sfmap estimates the hostname of an https server when given a pair of client and server IP addresses.} It can statistically estimate the hostname even when associating DNS queries are unobserved due to caching mechanisms, etc (challenge 2). Through extensive analysis using real packet traces , we demonstrate that the sfmap framework establishes good estimation accuracies and can outperform the state-of-the art technique called dn-hunter, [2]. {\color{red} We also identify the optimized setting of the sfmap framework. The experiment results suggest that the success of the sfmap lies in the fact that it can complement incomplete DNS information by leveraging the graph structure. To cope with large-scale measurement data, we introduce techniques to make the sfmap framework scalable. We validate the effectiveness of the approach using large-scale traffic data collected at a gateway point of internet access links.} The remainder of this paper is organized as follows: section2 summarizes the related work. [...]\\
\bottomrule
\end{tabularx}
\end{center}
\caption{\label{fig:summ_ex} An example of a document with summary statements highlighted in context.
}
\end{table}

In more detail, our contributions are as follows:
\begin{itemize}[noitemsep]
\item{We introduce a new dataset for summarisation of scientific publications consisting of over 10k documents}
\item{Following the approach of \cite{hermann2015teaching} in the news domain, we introduce a method, \textit{HighlightROUGE}, which can be used to automatically extend this dataset 
and show empirically that this improves summarisation performance}
\item{Taking inspiration from previous work in summarising scientific literature \citep{kupiec1995trainable, papers_citationSaggion2016}, we introduce a 
metric we use as a feature, \textit{AbstractROUGE}, which can be used to extract summaries by exploiting the abstract of a paper}
\item{We benchmark several neural as well traditional summarisation methods on the dataset and use simple features to model the global context of a summary statement, which contribute most to the overall score}
\item{We compare our best performing system to several well-established baseline methods, some of which use more elaborate methods to model the global context than we do, and show that our best performing model outperforms them on this extractive summarisation task by a considerable margin}
\item{We analyse to what degree different sections in scientific papers contribute to a summary}
\end{itemize}
We expect the research documented in this paper to be relevant beyond the document summarisation community, for other tasks in the space of automatically understand scientific publications, such as keyphrase extraction~\cite{kim-EtAl:2010:SemEval,sterckx2016supervised,augenstein2017scienceie,augenstein2017multitask}, semantic relation extraction~\cite{gupta-manning:2011:IJCNLP-2011,marsi-ozturk:2015:EMNLP} or topic classification of scientific articles~\cite{oseaghdha-teufel:2014:Coling}.

\section{Dataset and Problem Formulation}

\setlength{\tabcolsep}{0.3em}
\begin{table}[t]
\fontsize{10}{12}\selectfont
\begin{center}
\begin{tabular}{l c c c}
\toprule
& \bf \#documents & \bf \#instances \\
\midrule
CSPubSum Train & 10148 & 85490 \\ 
CSPubSumExt Train & 10148 & 263440 \\
CSPubSum Test & 150 & N/A \\ 
CSPubSumExt Test & 10148 & 131720 \\
\bottomrule
\end{tabular}
\end{center}
\caption{\label{tab:Dataset} The CSPubSum and CSPubSumExt datasets as described in Section~\ref{sec:training_data}. Instances are items of training data.
}
\end{table}

We release a novel dataset for extractive summarisation comprised of $10 148$ Computer Science publications.\footnote{The dataset along with the code is available here: \url{https://github.com/EdCo95/scientific-paper-summarisation}} Publications were obtained from ScienceDirect, where publications are grouped into 27 domains, Computer Science being one of them. As such, the dataset could easily be extended to more domains. 
An example document is shown in Table \ref{fig:summ_ex}.
Each paper in this dataset is guaranteed to have a title, abstract, author written highlight statements and author defined keywords. The highlight statements are sentences that should effectively convey the main takeaway of each paper and are a good gold summary, while the keyphrases are the key topics of the paper.
Both abstract and highlights can be thought of as a summary of a paper. Since highlight statements, unlike sentences in the abstract, generally do not have dependencies between them, we opt to use those as gold summary statements for developing our summarisation models, following \newcite{hermann2015teaching,nallapati2016abstractive} in their approaches to news summarisation.

\subsection{Problem Formulation}

As shown by \newcite{Cao2015}, sentences can be good summaries even when taken out of the context of the surrounding sentences. Most of the highlights have this characteristic, not relying on any previous or subsequent sentences to make sense. Consequently, we frame the extractive summarisation task here as a binary sentence classification task, where we assign each sentence in a document a label $y \in {0,1}$. Our training data is therefore a list of sentences, sentence features to encode context and a label all stored in a randomly ordered list.

\subsection{Creation of the Training and Testing Data}
\label{sec:training_data}

We used the 10k papers to create two different datasets: \textit{CSPubSum} and \textit{CSPubSumExt} where CSPubSumExt is CSPubSum extended with HighlightROUGE. The number of training items for each is given in Table~\ref{tab:Dataset}.

\paragraph{CSPubSum} This dataset's positive examples are the highlight statements of each paper. There are an equal number of negative examples which are sampled randomly from the bottom 10\% of sentences which are the worst summaries for their paper, measured with ROUGE-L (see below), resulting in $85490$ training instances. CSPubSum Test is formed of 150 full papers rather than a randomly ordered list of training sentences. These are used to measure the summary quality of each summariser, not the accuracy of the trained models.


\paragraph{CSPubSumExt} The CSPubSum dataset has two drawbacks: 1) it is an order of magnitude behind comparable large summarisation datasets~\cite{hermann2015teaching,nallapati2016abstractive}; 2) it does not have labels for sentences in the context of the main body of the paper. We generate additional training examples for each paper with \textit{HighlightROUGE} (see next section), which finds sentences that are similar to the highlights. This results in 263k instances for CSPubSumExt Train and 132k instances for CSPubSumExt Test. CSPubSumExt Test is used to test the accuracy of trained models. The trained models are then used in summarisers whose quality is tested on CSPubSum Test with the ROUGE-L metric (see below).





\section{ROUGE Metrics}\label{sec:ROUGE}

ROUGE metrics are evaluation metrics for summarisation which correspond well to human judgements of good summaries \citep{Lin2004}. We elect to use ROUGE-L, inline with other research into summarisation of scientific articles \citep{used_rouge_l_Cohan2015, dataJaidka2016}.


\subsection{HighlightROUGE}
\label{sec:highlight_rouge}
HighlightROUGE is a method 
used to generate additional training data for this dataset, using a similar approach to \cite{hermann2015teaching}. As input it takes a gold summary and body of text
and finds the sentences within that text which give the best ROUGE-L 
score in relation to the highlights, like an oracle summariser would do. These sentences represent the ideal sentences to extract from each paper for an extractive summary.




We select the top 20 sentences which give the highest ROUGE-L score with the highlights for each paper as positive instances and combine these with the highlights to give the positive examples for each paper. An equal number of negative instances are sampled from the lowest scored sentences to match.

When generating data using HighlightROUGE, no sentences from the abstracts of any papers were included as training examples. This is because the abstract is already a summary
; our goal is to extract salient sentences from the main paper to supplement the abstract, not from the preexisting summary.

\subsection{AbstractROUGE}
\label{sec:abs_rouge}
AbstractROUGE is used as a feature for summarisation. It is a metric presented by this work which exploits the known structure of a paper by making use of the abstract, a preexisting summary. 
The idea of AbstractROUGE is that sentences which are good summaries of the abstract are also likely to be good summaries of the highlights. The AbstractROUGE score of a sentence is simply the ROUGE-L score of that sentence and the abstract. The intuition of comparing sentences to the abstract is one often used in summarising scientific literature, e.g. \cite{papers_citationSaggion2016, kupiec1995trainable}, however these authors generally encode sentences and abstract as TF-IDF vectors, then compare them, rather than directly comparing them with an evaluation metric. While this may seem somewhat like cheating, all scientific papers are guaranteed to have an abstract so it makes sense to exploit it as much as possible. 


\section{Method}



We encode each sentence in two different ways: as their mean averaged word embeddings and as their Recurrent Neural Network (RNN) encoding.

\subsection{Summariser Features}
\label{sec:handcrafted_feats}
 As the sentences in our dataset are randomly ordered, there is no readily available context for each sentence from surrounding sentences (taking this into account is a potential future development). To provide local and global context, a set of 8 features are used for each sentence which are described below. These contextual features contribute to achieving the best performances. Some recent work in summarisation uses as many as 30 features \citep{modernfeaturesDlikman2016, modernfeaturesLitvak2016}. We choose only a minimal set of features to focus more on learning from raw data than on feature engineering, although this could potentially further improve results.

\paragraph{AbstractROUGE}
A new metric presented by this work, described in Section \ref{sec:abs_rouge}.

\paragraph{Location}
Authors such as \newcite{papersKavila2015} only chose summary sentences from the Abstract, Introduction or Conclusion, thinking these more salient to summaries; and we show that certain sections within a paper are more relevant to summaries than others (see Section~\ref{sec:finding_relevant_section}). Therefore we assign sentences an integer location for 7 different sections: Highlight, Abstract, Introduction, Results / Discussion / Analysis, Method, Conclusion, all else.\footnote{based on a small manually created gazetteer of alternative names} Location features have been used in other ways in previous work on summarising scientific literature; \newcite{Visser2009} extract sentence location features based on the headings they occurred beneath while \newcite{Teufel2002} divide the paper into 20 equal parts and assign each sentence a location based on which segment it occurred in - an attempt to capture distinct zones of the paper. 

\paragraph{Numeric Count}
is the number of numbers in a sentence, based on the intuition that sentences containing heavy maths are unlikely to be good summaries when taken out of context.

\paragraph{Title Score}
In \newcite{Visser2009} and \newcite{Teufel2002}'s work on summarising scientific papers, one of the features used is Title Score. Our feature differs slightly from \newcite{Visser2009} in that we only use the main paper title whereas \newcite{Visser2009} use all section headings. To calculate this feature, the non-stopwords that each sentence contains which overlap with the title of the paper are counted. 

\paragraph{Keyphrase Score}
Authors such as \newcite{SparckJones2007} refer to the keyphrase score as a useful summarisation feature. The feature uses author defined keywords and counts how many of these keywords a sentence contains, the idea being that important sentences will contain more keywords.

\paragraph{TF-IDF}
Term Frequency, Inverse Document Frequency (TF-IDF) is a measure of how relevant a word is to a document \citep{Ramos2003}. It takes into account the frequency of a word in the current document and the frequency of that word in a background corpus of documents; if a word is frequent in a document but infrequent in a corpus it is likely to be important to that document. TF-IDF was calculated for each word in the sentence, and averaged over the sentence to give a TF-IDF score for the sentence. Stopwords were ignored.

\paragraph{Document TF-IDF}
Document TF-IDF calculates the same metric as TF-IDF, but uses the count of words in a sentence as the term frequency and count of words in the rest of the paper as the background corpus. This gives a representation of how important a word is in a sentence in relation to the rest of the document.

\paragraph{Sentence Length}
Teufel et al. \citeyearpar{Teufel2002} created a binary feature for if a sentence was longer than a threshold. We simply include the length of the sentence as a feature; an attempt to capture the intuition that short sentences are very unlikely to be good summaries because they cannot possibly convey as much information as longer sentences.

\subsection{Summariser Architectures}
\label{sec:summ_architectures}
Models detailed in this section could take any combination of four possible inputs, and are named accordingly:
\begin{itemize}
\item{S: The sentence encoded with an RNN.}
\item{A: a vector representation of the abstract of a paper, created by averaging the word vectors of every non-stopword word in the abstract. Since an abstract is already a summary, this gives a good sense of relevance. It is another way of taking the abstract into consideration by using neural methods as opposed to a feature. A future development is to encode this with an RNN.}
\item{F: the 8 features listed in Section \ref{sec:handcrafted_feats}.}
\item{Word2Vec: the sentence represented by taking the average of every non-stopword word vector in the sentence.}
\end{itemize}
Models containing ``Net'' use a neural network with one or multiple hidden layers. Models ending with ``Ens'' use an ensemble. All non-linearity functions are Rectified Linear Units (ReLUs), chosen for their faster training time and recent popularity \cite{Krizhevsky2012}.

\paragraph{Single Feature Models}
The simplest class of summarisers use a single feature from Section \ref{sec:handcrafted_feats} (Sentence Length, Numeric Count and Section are excluded due to lack of granularity when sorting by these).

\paragraph{Features Only: FNet} 
\label{sec:fnet}
A single layer neural net to classify each sentence based on all of the 8 features given in Section~\ref{sec:handcrafted_feats}. A future development is to try this with other classification algorithms.

\paragraph{Word Vector Models: Word2Vec and Word2VecAF}
Both single layer networks. Word2Vec takes as input the sentence represented as an averaged word vector of 100 numbers.\footnote{Word embeddings are obtained by training a Word2Vec skip-gram model on the 10000 papers with dimensionality 100, minimum word count 5, a context window of 20 words and downsample setting of 0.001} Word2VecAF takes the sentence average vector, abstract average vector and handcrafted features, giving a 208-dimensional vector for classification.

\paragraph{LSTM-RNN Method: SNet}
\label{sec:lstm}
Takes as input the ordered words of the sentence represented as 100-dimensional vectors and feeds them through a bi-directional RNN with Long-Short Term Memory (LSTM, ~\newcite{Hochreiter1997}) cells, with 128 hidden units and dropout to prevent overfitting. Dropout probability was set to 0.5 which is thought to be near optimal for many tasks \citep{Srivastava2014}. Output from the forwards and backwards LSTMs is concatenated and projected into two classes.\footnote{The model is trained until loss convergence on a small dev set}

\paragraph{LSTM and Features: SFNet}
SFNet processes the sentence with an LSTM as in the previous paragraph and passes the output through a fully connected layer with dropout. The handcrafted features are treated as separate inputs to the network and are passed through a fully connected layer. The outputs of the LSTM and features hidden layer are then concatenated and projected into two classes.

\begin{figure}[t]
\centering
\includegraphics[width=\linewidth]{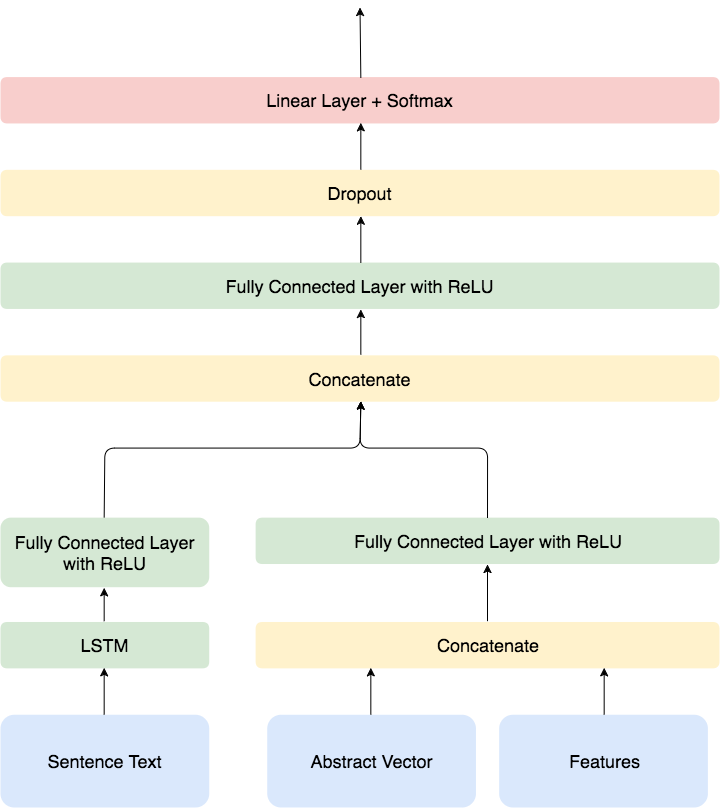}
\caption{SAFNet Architecture}
\label{fig:summnet}
\end{figure}

\paragraph{SAFNet}
SAFNet, shown in Figure \ref{fig:summnet} is the most involved architecture presented in this paper, which further to SFNet also encodes the abstract. 

\paragraph{Ensemble Methods: SAF+F and S+F Ensemblers} The two ensemble methods use a weighted average of the output of two different models:

$$
p_\text{summary} = \frac{S_1 (1 - C) + S_2 (1 + C)}{2}
$$

Where $S_1$ is the output of the first summariser, $S_2$ is the output of the second and $C$ is a hyperparameter. SAF+F Ensembler uses SAFNet as as $S_1$ and FNet as $S_2$. 
S+F Ensembler uses SNet as $S_1$ and FNet as $S_2$.

\section{Results and Analysis}

\subsection{Most Relevant Sections to a Summary}
\label{sec:finding_relevant_section}

A straight-forward heuristic way of obtaining a summary automatically would be to identify which sections of a paper generally represent good summaries and take those sections as a summary of the paper. This is precisely what \newcite{papersKavila2015} do, constructing summaries only from the Abstract, Introduction and Conclusion. This approach works from the intuition that certain sections are more relevant to summaries.

To understand how much each section contributes to a gold summary, 
we compute the ROUGE-L score of each sentence compared to the gold summary and average sentence-level ROUGE-L scores by section. 
ROUGE-type metrics are not the only metrics which we can use to determine how relevant a sentence is to a summary. Throughout the data, there are approximately 2000 occurrences of authors directly copying sentences from within the main text to use as highlight statements. By recording from which sections of the paper these sentences came, we can determine from which sections authors most frequently copy sentences to the highlights, so may be the most relevant to a summary. This is referred to as the \textit{Copy/Paste Score} in this paper. 

\begin{figure}[t]
\centering
\includegraphics[width=\linewidth]{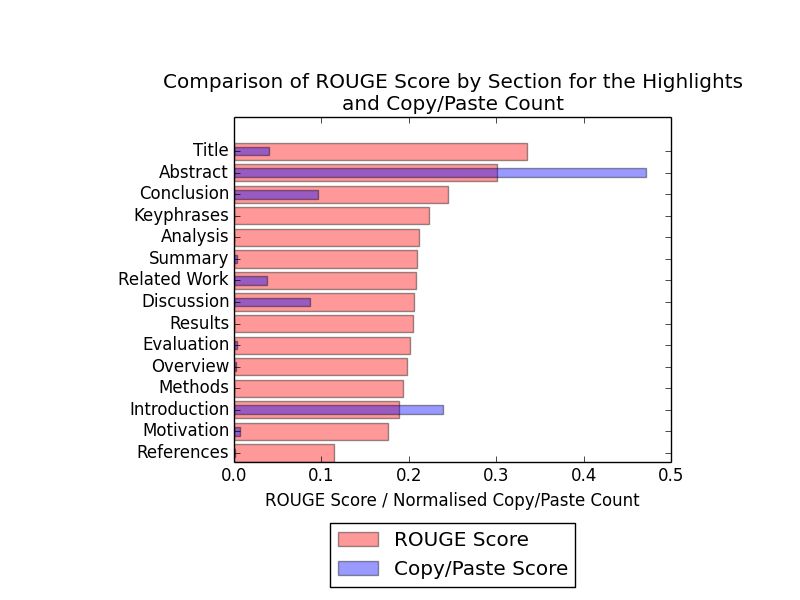}
\caption{Comparison of the average ROUGE scores for each section and the Normalised Copy/Paste score for each section, as detailed in Section \ref{sec:finding_relevant_section}. The wider bars in ascending order are the ROUGE scores for each section, and the thinner overlaid bars are the Copy/Paste count.}
\label{fig:rouge_by_section}
\end{figure}

Figure~\ref{fig:rouge_by_section} shows the average ROUGE score for each section over all papers, and the normalised Copy/Paste score. 
The title has the highest ROUGE score in relation to the gold summary, which is intuitive as the aim of a title is to convey information about the research in a single line. 

A surprising result is that the introduction has the third-lowest ROUGE score in relation to the highlights. Our hypothesis was that the introduction would be ranked highest after the abstract and title because it is designed to give the reader a basic background of the problem. Indeed, the introduction has the second highest Copy/Paste score of all sections. The reason the introduction has a low ROUGE score but high Copy/Paste score is likely due to its length. The introduction tends to be longer (average length of 72.1 sentences) than other sections, but still of a relatively simple level compared to the method (average length of 41.6 sentences), thus has more potential sentences for an author to use in highlights, giving the high Copy/Paste score. However it would also have more sentences which are not good summaries and thus reduce the overall average ROUGE score of the introduction. 

Hence, although some sections are slightly more likely to contain good summary sentences, and assuming that we do not take summary sentences from the abstract which is already a summary, then Figure \ref{fig:rouge_by_section} suggests that there is no definitive section from which summary sentences should be extracted.

\begin{figure}[t]
\centering
\includegraphics[width=\linewidth]{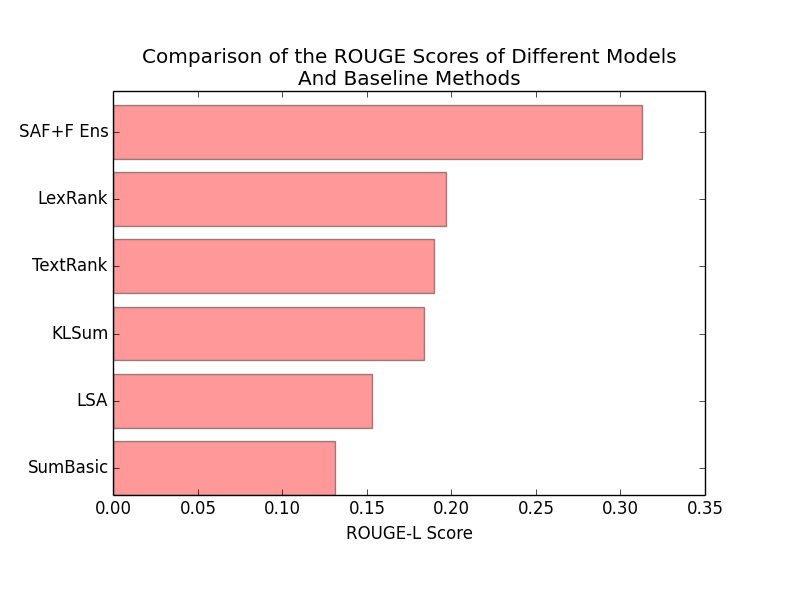}
\caption{Comparison of the best performing model and several baselines by ROUGE-L score on CSPubSum Test.}
\label{fig:model_comparison_baselines}
\end{figure}

\begin{figure}[t]
\centering
\includegraphics[width=\linewidth]{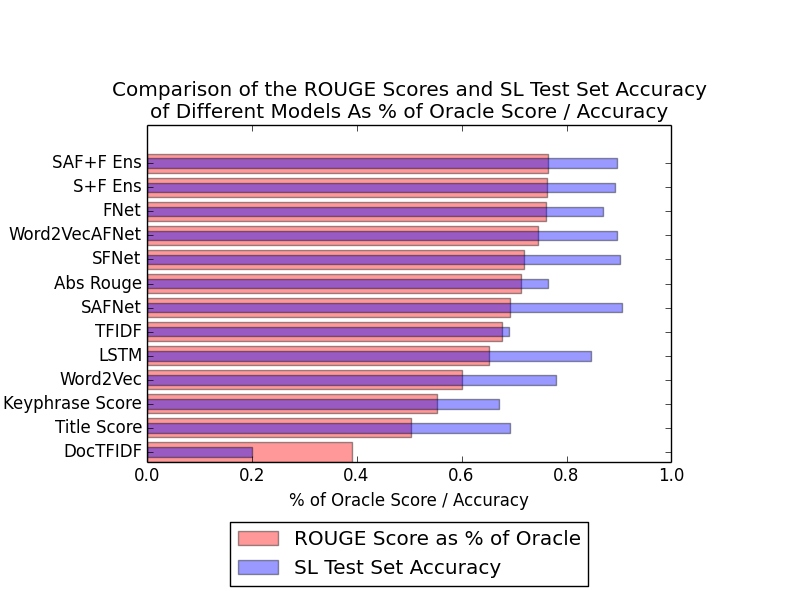}
\caption{Comparison of the accuracy of each model on CSPubSumExt Test and ROUGE-L score on CSPubSum Test. ROUGE Scores are given as a percentage of the Oracle Summariser score which is the highest score achievable for an extractive summariser on each of the papers. The wider bars in ascending order are the ROUGE scores. There is a statistically significant difference between the performance of the top four summarisers and the 5th highest scoring one (unpaired t-test, p=0.0139).}
\label{fig:model_comparison}
\end{figure} 

\subsection{Comparison of Model Performance and Error Analysis}
Figure~\ref{fig:model_comparison_baselines} shows comparisons of the best model we developed to well-established external baseline methods. Our model can be seen to significantly outperform these methods, including graph-based methods which take account of global context: LexRank \citep{lexrankRadev2004} and TextRank \citep{textrankMihalcea2004}; probabilistic methods in KLSum (KL divergence summariser, \newcite{klsumHaghighi2009}); methods based on singular value decomposition with LSA (latent semantic analysis, \newcite{lsaSteinberger2004}); and simple methods based on counting in SumBasic \citep{sumbasicVanderwende2007}. This is an encouraging result showing that our methods that combine neural sentence encoding and simple features for representing the global context and positional information are very effective for modelling an extractive summarisation problem.

Figure~\ref{fig:model_comparison} shows the performance of all models developed in this work measured in terms of accuracy and ROUGE-L on CSPubSumExt Test and CSPubSum Test, respectively. Architectures which use a combination of sentence encoding and additional features performed best by both measures. The LSTM encoding on its own outperforms models based on averaged word embeddings by 6.7\% accuracy and 2.1 ROUGE points. This shows that the ordering of words in a sentence clearly makes a difference in deciding if that sentence is a summary sentence. This is a particularly interesting result as it shows that encoding a sentence with an RNN is superior to simple arithmetic, and provides an alternative to the recursive autoencoder proposed by  \cite{recursiveaeSocher2011} which performed worse than vector addition.

Another interesting result is that the highest accuracy on CSPubSumExt Test did not translate into the best ROUGE score on CSPubSum Test, although they are strongly correlated (Pearson correlation, R=$0.8738$). SAFNet achieved the highest accuracy on CSPubSumExt Test, however was worse than the AbstractROUGE Summariser on CSPubSum Test. This is most likely due to imperfections in the training data. A small fraction of sentences in the training data are mislabelled due to bad examples in the highlights which are exacerbated by the HighlightROUGE method. This leads to confusion for the summarisers capable of learning complex enough representations to classify the mislabelled data correctly. 

We manually examined 100 sentences from CSPubSumExt which were incorrectly classified by SAFNet. Out of those, 37 are mislabelled examples.
The primary cause of \textit{false positives} was lack of context (16 / 50 sentences) and long range dependency (10 / 50 sentences). Other important causes of false positives were mislabelled data (12 / 50 sentences) and a failure to recognise that mathematically intense sentences are not good summaries (7 / 50 sentences). Lack of context is when sentences require information from the sentences immediately before them to make sense. For example, the sentence ``The performance of such systems is commonly evaluated using the data in the matrix'' is classified as positive but does not make sense out of context as it is not clear what systems the sentence is referring to. 
A long-range dependency is when sentences refer to an entity that is described elsewhere in the paper, e.g. sentences referring to figures. 
These are more likely to be classified as summary statements when using models trained on automatically generated training data with HighlightROUGE, because they have a large overlap with the summary.

The primary cause of \textit{false negatives} was mislabelled data (25 / 50 sentences) and failure to recognise an entailment, observation or conclusion (20 / 50 sentences). Mislabelled data is usually caused by the presence of some sentences in the highlights which are of the form ``we set m=10 in this approach'', which are not clear without context. 
Such sentences should only be labelled as positive if they are part of multi-line summaries, which is difficult to determine automatically.

Failure to recognise an entailment, observation or conclusion is where a sentence has the form "entity X seems to have a very small effect on Y" for example, but the summariser has not learnt that this information is useful for a summary, possibly because it was occluded by mislabelled data.

SAFNet and SFNet achieve high accuracy on the automatically generated CSPubSumExt Test dataset, though a lower ROUGE score than other simpler methods such as FNet on CSPubSum Test. This is likely due to overfitting, which our simpler summarisation models are less prone to.
One option to solve this would be to manually improve the CSPubSumExt labels, the other to change the form of the training data.
Rather than using a randomised list of sentences and trying to learn objectively good summaries \cite{Cao2015}, each training example could be all the sentences in order from a paper, classified as either summary or not summary. The best summary sentences from within the paper would then be chosen using HighlightROUGE and used as training data, and an approach similar to \newcite{nallapati2016summarunner} could be used to read the whole paper sequentially and solve the issue of long-range dependencies and context.



The issue faced by SAFNet does not affect the ensemble methods so much as 
their predictions are weighted by a hyperparameter tuned with CSPubSum Test rather than CSPubSumExt. Ensemblers ensure good performance on both test sets as the two models are adapted to perform better on different examples.

In summary, our model performances show that: reading a sentence sequentially is superior to averaging its word vectors, simple features that model global context and positional information are very effective and a high accuracy on an automatically generated test set does not guarantee a high ROUGE-L score on a gold test set, although they are correlated. This is most likely caused by models overfitting data that has a small but significant proportion of mislabelled examples as a byproduct of being generated automatically.

\subsection{Effect of Using ROUGE-L to Generate More Data}
\begin{figure}[t]
\centering
\includegraphics[width=\linewidth]{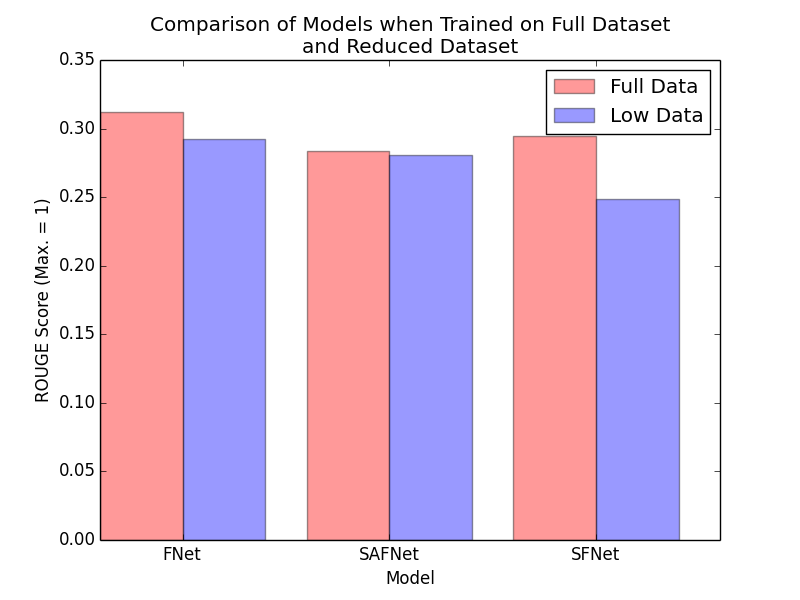}
\caption{Comparison of the ROUGE scores of FNet, SAFNet and SFNet when trained on CSPubSumExt Train (bars on the left) and CSPubSum Train (bars on the right) and .}
\label{fig:model_comparison_low_data}
\end{figure}

This work used a method similar to \newcite{hermann2015teaching} to generate extra training data (Section \ref{sec:highlight_rouge}). Figure \ref{fig:model_comparison_low_data} compares three models trained on CSPubSumExt Train and the same models trained on CSPubSum Train (the feature of which section the example appeared in was removed to do this). The FNet summariser and SFNet suffer statistically significant ($p=0.0147$ and $p<0.0001$) drops in performance from using the unexpanded dataset, although interestingly SAFNet does not, suggesting it is a more stable model than the other two. These drops in performance however show that using the method we have described to increase the amount of available training data does improves model performance for summarisation.

\subsection{Effect of the AbstractROUGE Metric on Summariser Performance}
\begin{figure}[t]
\centering
\includegraphics[width=\linewidth]{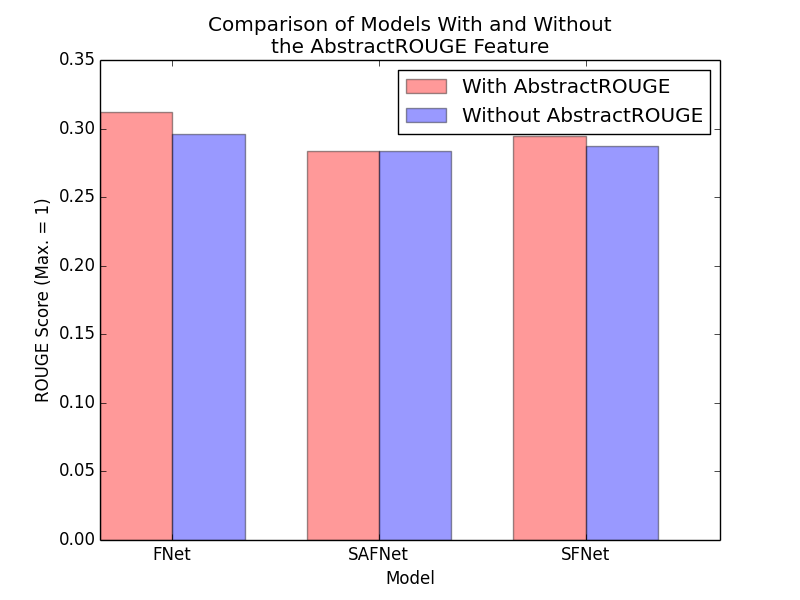}
\caption{Comparison of ROUGE scores of the Features Only, SAFNet and SFNet models when trained with (bars on the left) and without (bars on the right) AbstractROUGE, evaluated on CSPubSum Test. The FNet classifier suffers a statistically significant (p=0.0279) decrease in performance without the AbstractROUGE metric.}
\label{fig:model_comparison_no_abs_rouge}
\end{figure}

This work suggested use of the AbstractROUGE metric as a feature (Section \ref{sec:abs_rouge}). Figure \ref{fig:model_comparison_no_abs_rouge} compares the performance of 3 models trained with and without it. This shows two things: the AbstractROUGE metric does improve performance for summarisation techniques based only on feature engineering; and learning a representation of the sentence directly from the raw text as is done in SAFNet and SFNet as well as learning from features results in a far more stable model. This model is still able to make good predictions even if AbstractROUGE is not available for training, meaning the models need not rely on the presence of an abstract.

\section{Related Work}

\paragraph{Datasets}
Datasets for extractive summarisation often emerged as part of evaluation campaigns for summarisation of news, organised by the Document Understanding Conference (DUC), and the Text Analysis Conference (TAC).
DUC proposed single-document summarisation~\cite{harman2002duc}, whereas TAC datasets are for multi-document summarisation~\cite{dang2008overview,dang2009overview}. All of the datasets contain roughly 500 documents.


The largest summarisation dataset (1 million documents) to date is the DailyMail/CNN dataset~\cite{hermann2015teaching}, first used for single-document abstractive summarisation by~\cite{nallapati2016abstractive}, enabling research on data-intensive sequence encoding methods.


Existing datasets for summarisation of scientific documents of which we are aware are small. \newcite{kupiec1995trainable} used only 21 publications and CL-SciSumm 2017\footnote{\url{http://wing.comp.nus.edu.sg/cl-scisumm2017/}} contains 30 publications. \newcite{dataRonzano2016} used a set of 40 papers, \newcite{kupiec1995trainable} used 21 and \newcite{Visser2009} used only 9 papers. The largest known scientific paper dataset was used by \newcite{Teufel2002} who used a subset of 80 papers from a larger corpus of 260 articles.

The dataset we introduce in this paper is, to our knowledge, the only large dataset for extractive summarisation of scientific publications. The size of the dataset enables training of data-intensive neural methods and also offers exciting research challenges centered around how to encode very large documents.

\paragraph{Extractive Summarisation Methods}
Early work on extractive summarisation focuses exclusively on easy to compute statistics, e.g. word frequency~\cite{luhn1958automatic}, location in the document~\cite{baxendale1958machine}, and TF-IDF~\cite{salton1996automatic}.
Supervised learning methods which classify sentences in a document binarily as summary sentences or not soon became popular~\cite{kupiec1995trainable}. Exploration of more cues such as sentence position~\cite{yang-bao-nenkova:2017:EACLshort}, sentence length~\cite{radev2004mead}, words in the title, presence of proper nouns, word frequency \cite{nenkova2006compositional} and event cues~\cite{filatova2004event} followed.

Recent approaches to extractive summarisation have mostly focused on neural approaches, based on bag of word embeddings approaches~\cite{kobayashi2015summarization,yogatama2015extractive} or encoding whole documents with CNNs and/or RNNs~\cite{cheng-lapata:2016:P16-1}. 

In our setting, since the documents are very large, it is computationally challenging to read a whole publication with a (possibly hierarchical) neural sequence encoder. 
In this work, we therefore opt to only encode the target sequence with an RNN and the global context with simpler features.
We leave fully neural approaches to encoding publications to future work.

\section{Conclusion}

In this paper, we have introduced a new dataset for summarisation of computer science publications, which is substantially larger than comparable existing datasets, by exploiting an existing resource. We showed the performance of several extractive summarisation models on the dataset that encode sentences, global context and position, which significantly outperform well-established summarisation methods. We introduced a new metric, AbstractROUGE, which we show increases summarisation performance. Finally, we show how the dataset can be extended automatically, which further increases performance.
Remaining challenges are to better model the global context of a summary statement and to better capture cross-sentence dependencies.

\section*{Acknowledgments}

This work was partly supported by Elsevier.


\bibliography{acl2017}
\bibliographystyle{acl_natbib}

\appendix


\end{document}